\documentclass{article}

\usepackage[preprint]{neurips_2026}
\usepackage[utf8]{inputenc}
\usepackage[T1]{fontenc}
\usepackage{hyperref}
\usepackage{url}
\usepackage{booktabs}
\usepackage{float}
\usepackage{amsfonts}
\usepackage{amsmath}
\usepackage{enumitem}
\usepackage{amssymb}
\usepackage{nicefrac}
\usepackage{microtype}
\usepackage{xcolor}
\usepackage{algorithm}
\usepackage{tcolorbox}

\usepackage{algpseudocode}
\usepackage{pifont}
\usepackage{graphicx}
\usepackage{multirow}
\definecolor{confirmedbg}{HTML}{EDF5EC}
\definecolor{confirmedframe}{HTML}{5A9B57}
\definecolor{partialbg}{HTML}{FDF4E7}
\definecolor{partialframe}{HTML}{C49A3C}

\newtcolorbox{confirmed}[1][]{%
  colback=confirmedbg,
  colframe=confirmedframe,
  boxrule=0.5pt,
  left=8pt, right=8pt, top=5pt, bottom=5pt,
  before upper={\textbf{\textcolor{confirmedframe}{#1}}\par\smallskip}
}

\newtcolorbox{partialresult}[1][]{%
  colback=partialbg,
  colframe=partialframe,
  boxrule=0.5pt,
  left=8pt, right=8pt, top=5pt, bottom=5pt,
  before upper={\textbf{\textcolor{partialframe}{#1}}\par\smallskip}
}

\newcommand{\wfr}{\texttt{word\_fragmentation\_rate}}
\newcommand{\ab}{AbsenceBench}
\title{The Text Uncanny Valley:\\ Non-Monotonic Performance Degradation \\in LLM Information Retrieval}

\author{%
  Zekai Tong\textsuperscript{1,$*$} \quad
  Ruiyao Xu\textsuperscript{2} \quad
  Aryan Shrivastava\textsuperscript{1} \\[4pt]
  \textbf{Chenhao Tan}\textsuperscript{1} \quad
  \textbf{Ari Holtzman}\textsuperscript{1} \\[6pt]
  \textsuperscript{1}University of Chicago \quad
  \textsuperscript{2}Northwestern University
}

\begin{document}

\maketitle
\footnotetext[1]{$^*$Corresponding author: \texttt{zekaitong@uchicago.edu}}

\begin{abstract}
Existing Large Language Model (LLM) benchmarks primarily focus on syntactically correct inputs, leaving a significant gap in evaluation on imperfect text. In this work, we study how word-boundary corruption affects how LLMs detect targeted information. 
By inserting whitespace characters within words to break them into fragments, LLMs' detection accuracy follows a
\textbf{U-shaped curve} with the increase in insertion rate.
We refer to this curve as the \textbf{Text Uncanny Valley}. To explain such observation, we propose a \emph{mode transition} hypothesis: LLMs operate in a
word-level mode for near-normal text and a character-level mode for
heavily fragmented text, with the valley marking the disordered transition where neither mode is effective. 
Four experiments and one analysis are consistent with this account: in-context learning fails to rescue
valley-bottom performance; regularizing the perturbation substantially
reduces the U-shape; a math reasoning task replicates the U-shape for
Gemini 3.0 Flash but not for stronger models, suggesting the effect is
attenuated when tasks rely less on exact lexical alignment; and tokenization
entropy peaks before the F1 minimum, consistent with a regime-conflict
interpretation.
These findings reveal a failure mode invisible to clean-text benchmarks
yet directly relevant to any deployment scenario involving noisy or
uncurated text inputs.
\end{abstract}

\begin{figure}[h!]
  \centering
  \includegraphics[width=0.68\linewidth]{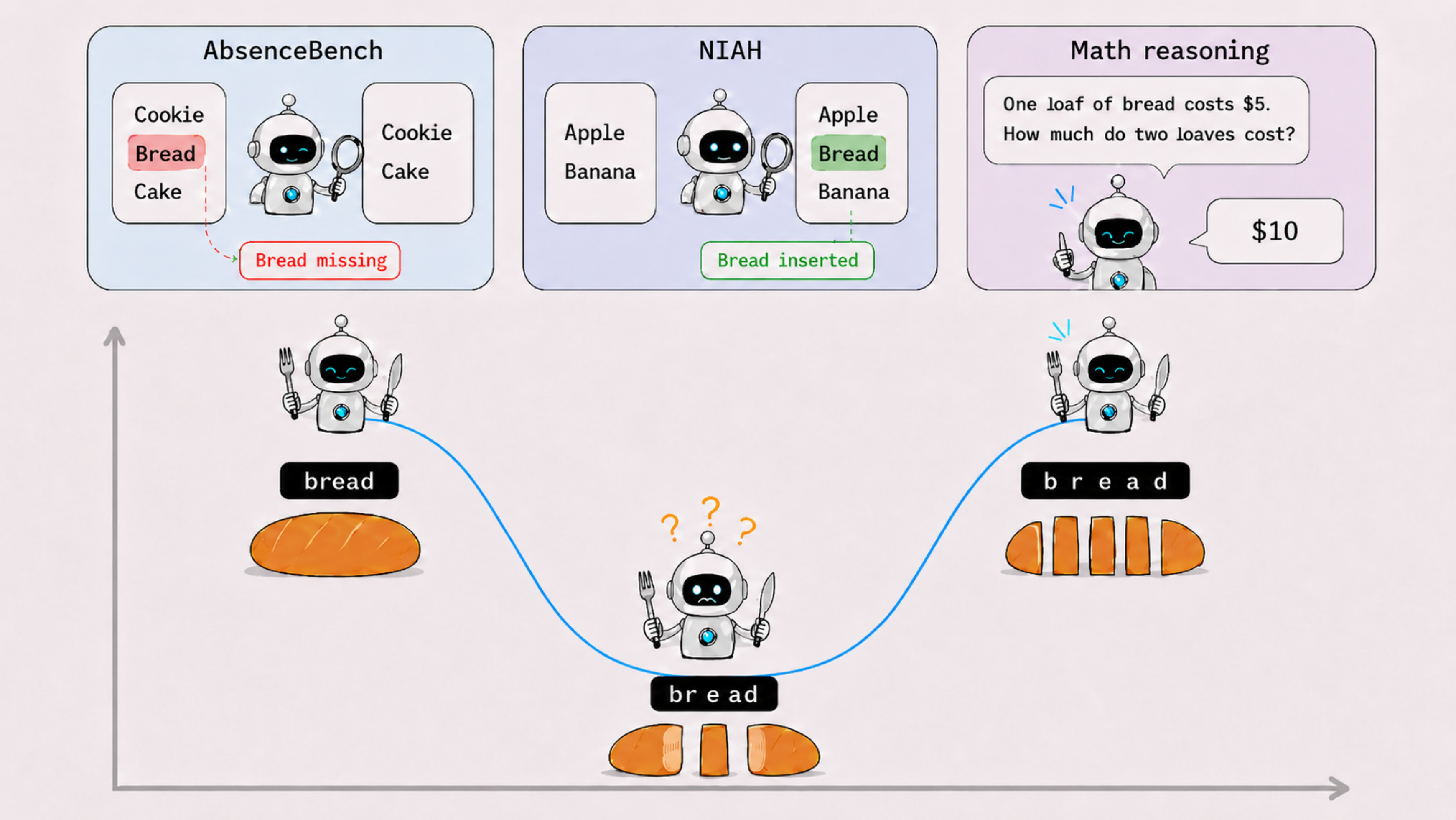}
  \caption{Bread slicing as an analogy for our text fragmentation method and
  the resulting Text Uncanny Valley phenomenon. We evaluate three tasks:
  AbsenceBench, NIAH-style task, and math reasoning. Image generated by ChatGPT.}
  \label{fig:bread}
\end{figure}

\section{Introduction}
\label{sec:intro}

Large language models (LLMs) are increasingly deployed in settings where input
text is far from clean. Documents extracted from PDFs via optical character
recognition (OCR), content copy-pasted from web pages, and outputs from
automated pipelines routinely contain formatting artifacts, including spurious
spaces inserted within words (e.g. \textit{international} $\to$ \textit{int er nation al}). Such corruption occurs routinely in any system that processes uncurated
text, but no existing benchmark measures its effect on LLM performance. Although recent work has begun to evaluate LLM robustness under noisy text
\citep{singh2024robustness, agrawal2025llmrobustness, piryani2025multiocr}, 
none examine how word-boundary corruption on tasks that demand \emph{exact}
span-level matching between two document versions, where tokenization
disruption has the most room to cause damage.

To address this gap, we study two complementary evaluation tasks: \textbf{absence detection}
\citep{fu2025absencebench}, where models identify deliberately omitted lines,
and a \textbf{NIAH-style} task \citep{kamradt2023niah}, where models identify
inserted lines. Both tasks require a model to align two document versions at the line
level, making them sensitive to any corruption that disrupts word
boundaries. To study this sensitivity in a controlled way, we insert spaces inside
words at a tunable rate \wfr{}~$\in [0,1]$, sweeping from clean text
(\wfr{}~$= 0$) to fully character-separated text (\wfr{}~$= 1$) across
eleven levels and three domains, including legal contracts, GitHub PRs, and
mathematical reasoning.

The most intuitive expectation is monotonic degradation: more corruption,
worse performance. What we observe instead is a \textbf{non-monotonic U-shaped} performance curve across the strongest model families: Micro-F1 degrades as
fragmentation increases, reaches a minimum at moderate rates, and then
partially \emph{recovers} at extreme fragmentation. We call this the \textbf{Text Uncanny Valley} \citep{mori1970uncanny, seyama2007uncanny}: text that is neither fully
natural nor uniformly fragmented leaves the model in a disordered
in-between state where performance is worst.

We evaluate eight model variants across three domains.
Three families, Gemini~3.0~Flash~\citep{gemini3flash}, GPT-5.2~\citep{gpt52systemcard}, and Claude~4.5~Haiku~\citep{claudehaiku45},
consistently exhibit the U-shape on both tasks and are the focus of subsequent analysis. That performance \emph{recovers} at high fragmentation, rather than
continuing to degrade, rules out a simple ``more noise, worse
performance'' account and points to a shift in how the model processes
text. We then formalize this as a \emph{two-regime hypothesis}: LLMs rely on
word-level processing for near-normal text and fall back to
character-level processing for heavily fragmented text, with the valley
marking the unstable transition between the two. Four experiments and a tokenization entropy analysis test this account.

Our contributions can be summarized into four folds:

\begin{itemize}[leftmargin=1.2em]
\item \textbf{Benchmark.}
        A controlled \wfr{} perturbation framework that sweeps
        word-boundary corruption from clean text to fully
        character-separated text across eleven levels, evaluated on
        two span-matching tasks and three domains.
\item \textbf{Phenomenon.} Three of eight model families, Gemini~3.0~Flash, GPT-5.2, and Claude~4.5~Haiku, exhibit a consistent U-shaped performance curve on both tasks, which we name the Text Uncanny Valley.
  \item \textbf{Explanation.}
        We propose a two-regime hypothesis and provide converging
        evidence:
        \ding{192}~ICL cannot rescue valley-bottom performance;
        \ding{193}~regularizing fragmentation reduces the U-shape while
        inter-word whitespace manipulation does not induce it;
        \ding{194}~tokenization entropy peaks before the F1 minimum;
        \ding{195}~a math reasoning control shows the effect is
        attenuated without cross-document alignment.
  \item \textbf{Implication.}
        Moderate word-boundary corruption, the level typical of OCR
        artifacts, is more damaging than extreme corruption, a failure
        mode invisible to clean-text benchmarks.
\end{itemize}

\section{Related Work}
\label{sec:related}

\textbf{Absence detection and NIAH.}
\citet{fu2025absencebench} establish that LLMs fail at absence detection on clean
text, achieving only around 70\% Micro-F1 despite near-perfect scores on NIAH
retrieval \citep{kamradt2023niah, hsieh2024ruler}.
We replicate and extend their evaluation to fragmented text.
\citet{jain2026sufficiency} and \citet{ming2025faitheval} study related but
distinct problems where missing information is semantically defined; neither
examines text-level corruption.

\textbf{LLM robustness to text perturbation.}
\citet{singh2024robustness} and \citet{zhuo2025charperturb} benchmark LLMs under
OCR errors, typos, and character-level perturbations; both find that models
tolerate mild noise but degrade under heavier corruption.
\citet{sclar2024prompt} show spacing variations can shift accuracy by up to 76
points.
\citet{pawar2025brokenwords} study tokenizer-induced word splitting.
Our key distinction: prior work reports monotonic degradation; we reveal a
\emph{non-monotonic} pattern where moderate fragmentation is worse than extreme,
via continuous rate sweeps that expose a recovery regime discrete evaluations miss.

\textbf{Other U-shaped curves in LLMs.}
\citet{liu2024lostinmiddle} document a U-shape as a function of information
\emph{position}; ours arises from intra-word fragmentation, a different axis
and mechanism.

\textbf{Tokenization and word boundaries.}
\citet{stop2026tokenizers} argue tokenization is a core modeling decision;
\citet{chai2024tokenizationfalling} show LLMs remain brittle to typographical
variations under BPE \citep{sennrich2016bpe}---a pattern we extend to
continuous intra-word space insertion with non-monotonic effects.

\textbf{In-context learning under noise.}
\citet{gao2024iclnoise} and \citet{jiao2026icldemo} show corrupted demonstrations
can hurt or mislead ICL, grounding our observation that ICL fails to rescue
valley-bottom performance.

\section{Method: Controlled Word-Boundary Fragmentation}
\label{sec:method}

\subsection{Perturbation Design}
\label{sec:method-perturb}

We perturb documents by inserting spaces within words.
For each \emph{contiguous non-space character sequence}, we iterate over its
\emph{intra-word slots} (positions between adjacent characters).
A sequence of $k$ characters has $k-1$ slots; \texttt{cat} has 2 slots, so
inserting at the first yields \texttt{c at} with \wfr{}~$= \tfrac{1}{2}$.
At each slot, we insert a space with probability $p \in [0, 1]$.
The degree of perturbation is measured by the
\textbf{word fragmentation rate} (\wfr{}), defined as:

\begin{equation}
  \wfr = \frac{\text{number of spaces actually inserted}}
              {\text{total number of intra-word slots}}
  \label{eq:wfr}
\end{equation}

\noindent
Because insertion is stochastic, the measured \wfr{} may differ slightly from
$p$, though in practice the difference is negligible.\footnote{Since the measured \wfr{} is nearly identical to $p$ across all experiments, we use $p$ to denote the fragmentation level but retain \wfr{} as the x-axis label for notational convenience.}
We sweep \wfr{} across eleven levels: $\{0.0, 0.1, 0.2, \ldots, 1.0\}$.
Appendix~\ref{app:frag-table} illustrates the effect on a legal document
excerpt at representative \wfr{} levels.

Two design properties make this perturbation suitable for controlled
experimentation:
\begin{itemize}[leftmargin=1.2em]
  \item \textbf{Continuous.} \wfr{} sweeps from 0.0 (clean input) to 1.0
        (every character pair separated by a space), providing a fine-grained
        view of how word-boundary integrity interacts with model capability.
  \item \textbf{Separable.} The target seed (which lines are omitted or inserted) and the
        fragmentation seed (where spaces are inserted) are independently
        controlled. Target masks are fixed across all fragmentation rates,
        ensuring that task difficulty does not covary with the perturbation
        level.
\end{itemize}

\subsection{Datasets}
\label{sec:method-data}

We evaluate on three domains, each with distinct textual characteristics.

\textbf{Legal documents.}
We use CUAD-QA \citep{hendrycks2021cuad}, expert-annotated commercial contracts
from the EDGAR filing system.
Dense natural-language prose makes these documents highly sensitive to
word-boundary disruption.

\textbf{GitHub pull requests.}
We use the same corpus as \citet{fu2025absencebench}: code diffs containing
long non-spaced identifier runs (e.g., \texttt{getUserByEmailAndPassword}),
making the text structurally dense before any perturbation.

\textbf{Mathematical reasoning (GSM8K).}
We draw problems from GSM8K \citep{cobbe2021gsm8k}, a dataset of 8,500
grade-school math word problems with step-by-step solutions, created by
OpenAI.\footnote{\url{https://huggingface.co/datasets/openai/gsm8k}}
To create documents of sufficient length to challenge modern LLMs, we
concatenate 80 problems per document, a volume approximating two weeks of student homework. Each question and its solution are treated as a sequence of lines split by periods and newlines.
The resulting document structure is illustrated in
Appendix~\ref{app:gsm8k-table}.

Due to the substantial computational cost of our evaluation protocol
(see Appendix~\ref{sec:limitations} for details), we sample 100 documents per
domain.

\subsection{Evaluation Tasks}
\label{sec:method-bench}

We evaluate word-boundary fragmentation on two tasks, both requiring exact
span-level matching between document versions.

\textbf{Absence detection.}
Following \citet{fu2025absencebench}, each instance provides both the original
document and a version from which lines have been deleted; the model must
identify exactly which lines are missing.
This task requires exact lexical alignment and is highly sensitive to
intra-word disruption.
The prompt template is in Appendix~\ref{app:prompt-table}.

\textbf{NIAH-style insertion detection.}
Each instance provides both the original document and a version into which
extra lines have been inserted; the model must identify exactly which lines
were added.
This adapts the classic NIAH setup \citep{kamradt2023niah} to require span
matching rather than simple retrieval, and uses the same Micro-F1 metric.
The prompt template and full experimental details are in Appendix~\ref{app:niah}.

Following \citet{fu2025absencebench}, both tasks are evaluated using Micro-F1 based on case-insensitive exact matching per line. This metric effectively penalizes false positives, providing a more rigorous assessment than recall alone.

Since we find that varying the omission/insertion rate does not substantially alter the U-shaped curve, we fix the omission/insertion rate at $p = 0.1$ for both tasks.
All models are queried via their respective APIs.
Model responses are parsed to extract the list of predicted target lines,
which are then compared against the ground-truth to compute
Micro-F1.\footnote{Full implementation and parsing details are provided in
Appendix~\ref{app:details}.}

\section{The Text Uncanny Valley}
\label{sec:valley}

\subsection{Main Finding: The U-shaped Curve and the Text Uncanny Valley}
\label{sec:valley-main}

We evaluate eight model variants:
DeepSeek-V3 \citep{deepseekv3}, Grok-4.1-Fast \citep{grok41}, Qwen3-Plus
\citep{qwen3}, Gemini 3.0 Flash (standard and thinking) \citep{gemini3flash},
GPT-5.2 (standard and thinking) \citep{gpt52systemcard}, and Claude 4.5 Haiku
\citep{claudehaiku45}.
Among these, only Gemini 3.0 Flash, GPT-5.2, and Claude 4.5 Haiku exhibit a
clear and consistent non-monotonic U-shaped curve across both tasks.
Grok-4.1-Fast shows largely monotonic decline without a discernible rebound.
Qwen3-Plus exhibits irregular curve shapes, generally poor performance, and
slow API responses with high cost, making it unsuitable for further analysis.
DeepSeek-V3 performs poorly throughout and frequently produces invalid outputs
at high fragmentation rates.
Full results for all eight variants are provided in
Appendix~\ref{app:all-models}.
The remainder of this paper focuses on the three model families that consistently
exhibit the Text Uncanny Valley.\footnote{GPT-5.2 returned API errors on a
small number of documents in some conditions due to context length limits;
affected documents were excluded from the corresponding averages.}

\paragraph{The Text Uncanny Valley.}
We name this non-monotonic curve the \textbf{Text Uncanny Valley}
\citep{mori1970uncanny, seyama2007uncanny}: just as human affinity drops for 
robots that are nearly but not perfectly realistic, LLMs enter a disordered 
transitional state when processing text that is neither fully natural nor 
uniformly fragmented.
The Micro-F1 drop predominantly reflects genuine detection failures rather than
a formatting artifact.

\begin{figure}[t]
  \centering
  \includegraphics[width=0.33\linewidth]{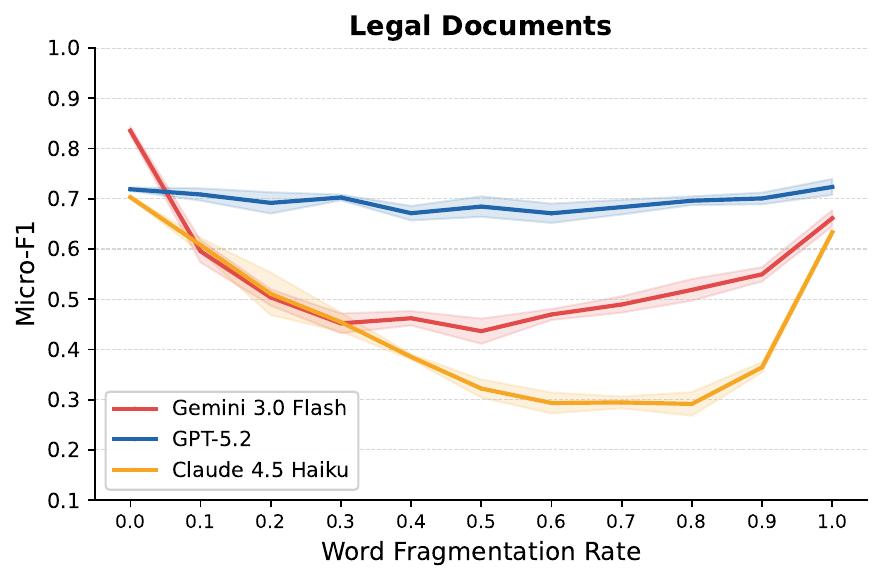}%
  \includegraphics[width=0.33\linewidth]{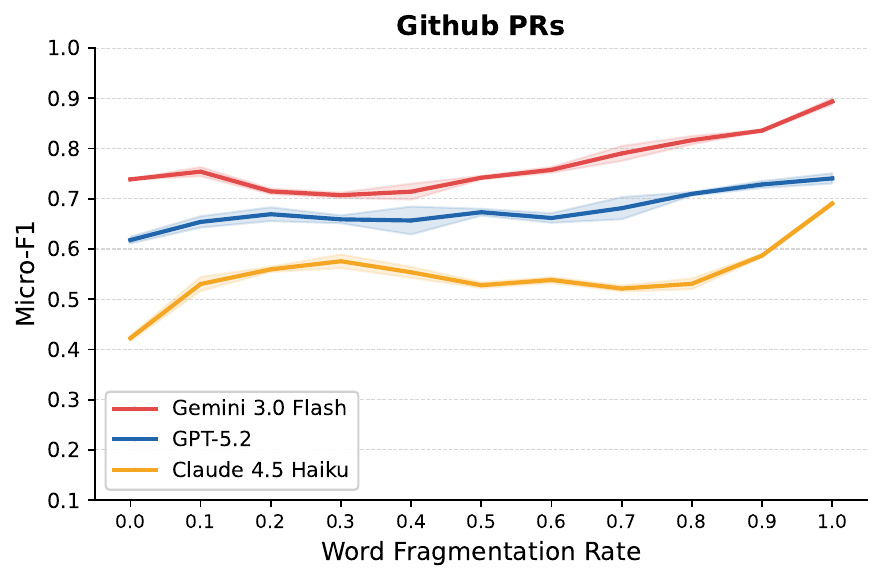}%
  \includegraphics[width=0.33\linewidth]{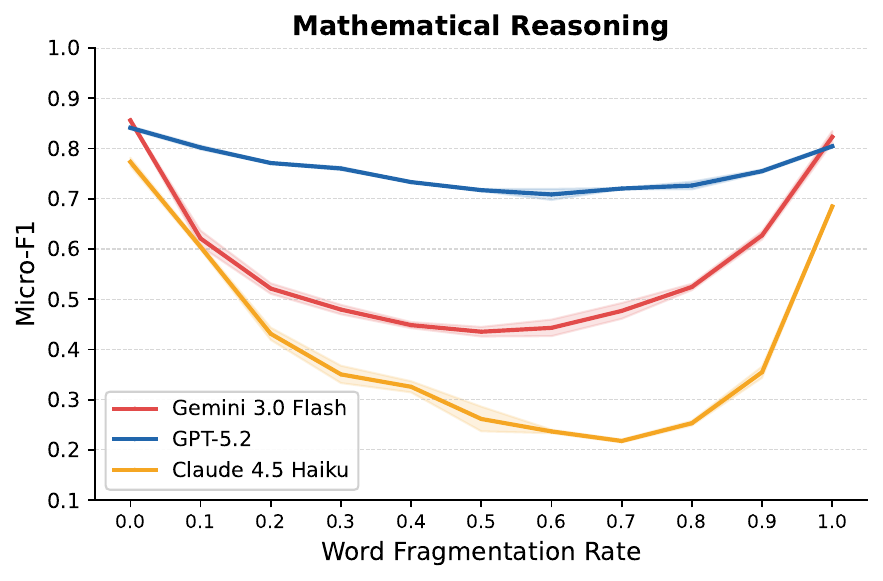}
  \caption{
    \textbf{The Text Uncanny Valley}: Micro-F1 vs.\ \wfr{} for absence
    detection on Legal Documents (left), GitHub PRs (centre), and Mathematical Reasoning
    (right). Shaded bands: $\pm$1 std over three fragmentation seeds.
  }
  \label{fig:main}
\end{figure}

\begin{figure}[t]
  \centering
  \includegraphics[width=0.33\linewidth]{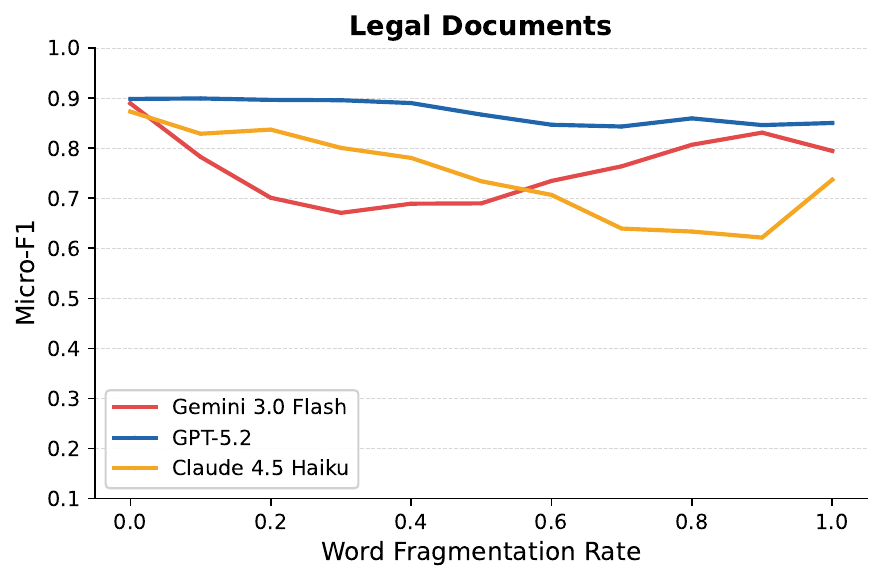}%
  \includegraphics[width=0.33\linewidth]{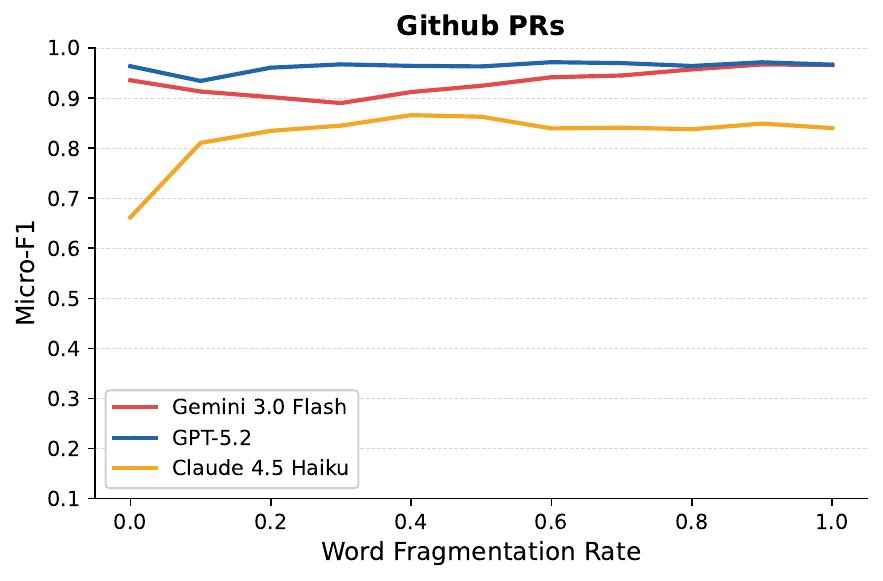}%
  \includegraphics[width=0.33\linewidth]{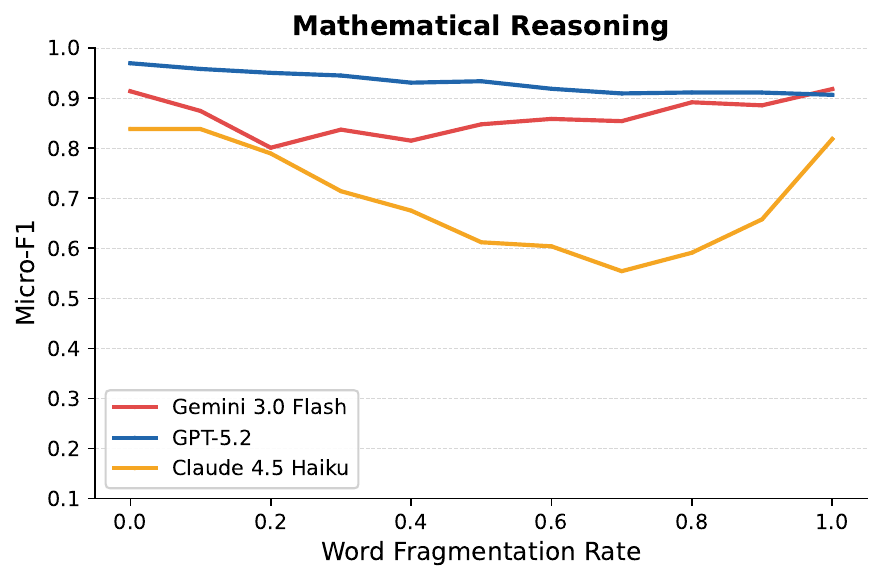}
  \caption{
    NIAH-style insertion detection under fragmentation.
    Results for the same domains and models as Figure~\ref{fig:main}. While the U-shaped \textbf{Text Uncanny Valley} remains present, the effect is less pronounced than in the absence detection task.
  }
  \label{fig:niah-main}
\end{figure}

The U-shaped curve appears in both tasks, but is more pronounced in absence
detection than in NIAH-style insertion detection.
Since the effect is clearest in absence detection, the follow-up experiments
in Section~\ref{sec:experiments} use absence detection as the primary testbed.

\subsection{Variation Across Domains and Models}
\label{sec:valley-variation}

Although all three models exhibit a U shape across all domains in absence detection, and in most cases for NIAH, the curve varies substantially with both the domain and the model family.

\textbf{GitHub PRs vs.\ natural-language domains.}
On Legal Documents and Mathematical Reasoning, all three models follow the expected
U-shape from the clean baseline in absence detection.
On GitHub PRs, however, all three models show an initial \emph{rise} in
Micro-F1 at low \wfr{} before turning downward.
GitHub PR diffs are already dense with long non-spaced identifier runs, and
a small number of inserted spaces may help by breaking these into shorter,
more parseable units before heavier fragmentation overwhelms this benefit.
This initial rise pattern appears in absence detection and for Claude 4.5 Haiku in NIAH, 
suggesting it reflects domain structure.

\textbf{Legal Documents vs.\ Mathematical Reasoning.}
In most cases, the valley is noticeably deeper on Mathematical Reasoning than on Legal Documents,
suggesting the effect is amplified when lines share strict logical
dependencies.\footnote{Establishing a causal link would require controlled
experiments varying dependency structure, which we leave to future work.}

\textbf{Model comparison.}
Across both tasks, GPT-5.2 consistently produces the flattest curve
and achieves the highest absolute Micro-F1 in most settings, suggesting greater
robustness to word-boundary fragmentation.
Claude 4.5 Haiku shows the deepest valley in absence detection, with its valley
bottom noticeably right-shifted relative to the other two models; this pattern
is less prominent in NIAH, possibly because the insertion detection task is
overall less sensitive to the transition zone.
Gemini 3.0 Flash falls between the two in most cases.

\subsection{Thinking Mode Does not Consistently Help}
\label{sec:thinking}

\citet{fu2025absencebench} report a 7.9\% thinking-mode boost on clean text.
We test whether this extends to fragmented text, evaluating Gemini 3.0 Flash
and GPT-5.2 on Legal Documents under absence detection (which shows the stronger
U-shape; NIAH evaluation is left to future work).

\begin{figure}[t]
  \centering
  \includegraphics[width=0.33\linewidth]{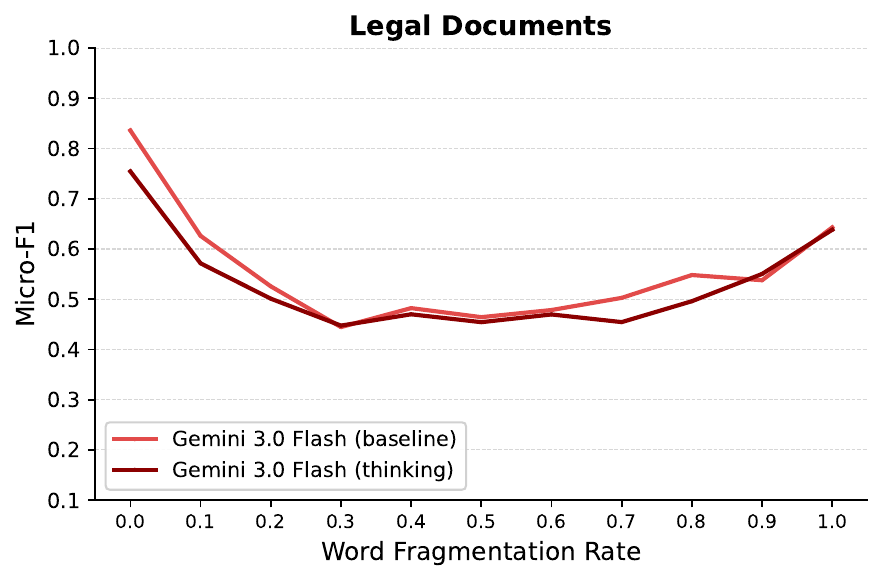}%
  \hspace{0.02\linewidth}%
  \includegraphics[width=0.33\linewidth]{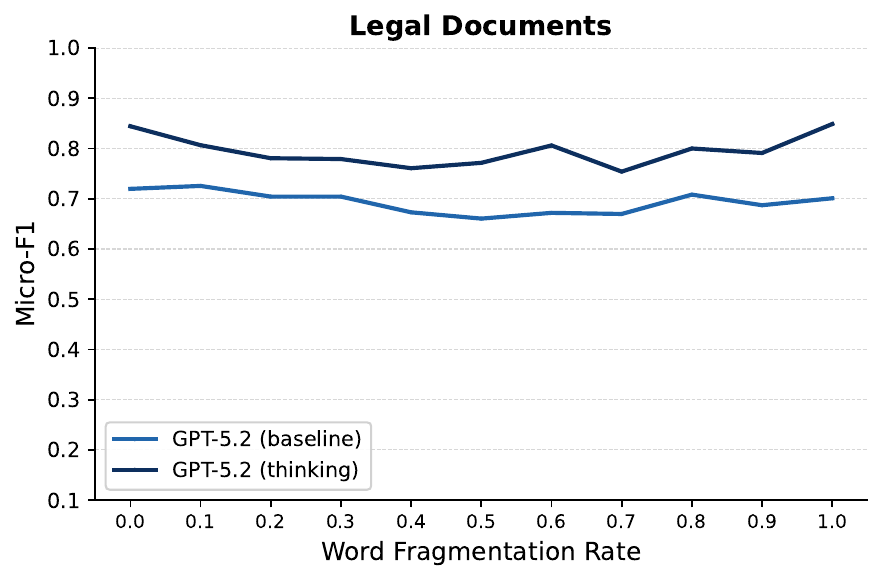}
  \caption{
    Comparison of standard and thinking mode on legal documents for
    Gemini 3.0 Flash (left) and GPT-5.2 (right).
  }
  \label{fig:thinking}
\end{figure}

Figure~\ref{fig:thinking} shows the results.
The U-shaped curve persists in both thinking-mode variants, but the effect
differs markedly between models.
Gemini 3.0 Flash thinking mode fails to improve performance in most conditions
and sometimes performs worse than the baseline.
GPT-5.2 thinking mode consistently improves Micro-F1 across fragmentation rates,
though it does not eliminate the valley.
Given the substantial increase in latency and cost, the cost-benefit tradeoff
of thinking mode is unfavorable in this setting, at least for Gemini.\footnote{We
do not extend thinking-mode evaluation beyond Gemini 3.0 Flash and GPT-5.2 on
Legal Documents; broader evaluation is a direction for future work.}

\section{Hypothesis: Two Processing Modes}
\label{sec:hypothesis}

The recovery at high fragmentation rules out a simple ``more noise,
worse output'' account: if fragmentation only destroyed information,
performance would decline monotonically.
That performance \emph{rebounds} means the model at \wfr{}~$=1.0$
operates in a qualitatively different state than at \wfr{}~$=0.5$. What changes between these two points is the tokenizer's output. At low \wfr{}, subword tokenizers produce familiar multi-character
tokens and the model reasons over words as coherent units.
At \wfr{}~$=1.0$, every character is space-separated, collapsing
the vocabulary to near-uniform single-character tokens; the model
falls back to character-level pattern matching, a regime where
``word recovery'' has been independently
documented~\citep{yang2026wordrecovery}.
Both extremes are internally consistent.
The problem arises at intermediate \wfr{}, where words are inconsistently affected: while some remain intact, others are partially or fully shattered into irregular fragments;
the model cannot commit to either processing strategy.

\begin{tcolorbox}[
  colback=cyan!8, colframe=cyan!50!blue!40,
  title={\textbf{Two-Regime Hypothesis}},
  fonttitle=\small, boxrule=0.5pt, arc=2pt
]
LLMs process text in two regimes:
\begin{enumerate}[leftmargin=1.5em, itemsep=2pt, topsep=2pt]
  \item \textbf{Word-level regime} (\wfr{}~$\approx 0$):
    subword tokenization is stable; the model aligns and compares
    documents through lexical units.
  \item \textbf{Character-level regime} (\wfr{}~$\approx 1$):
    tokenization collapses to single characters; the model effectively operates on character patterns to recover semantics.
\end{enumerate}
The \emph{Text Uncanny Valley} is the transition zone where
tokenization is neither coherently subword nor uniformly
character-level, and neither strategy is effective.
\end{tcolorbox}

This hypothesis yields testable predictions:
if the bottleneck is tokenization level, prompt level interventions such as
in-context learning should not resolve it;
if it is the \emph{irregularity} of fragmentation that triggers the valley,
then more regular fragmentation patterns should substantially reduce or eliminate
it;
and if the valley marks a transition between two tokenization regimes, then
tokenization heterogeneity should peak near the same region as the F1 minimum.

\label{sec:experiments}

\subsection{ICL Cannot Rescue the Valley}
\label{sec:exp-icl}

\textbf{Setup.} If the valley arises from a lack of task strategy, providing few-shot examples
of correctly handled fragmented text should help.
We prepend 3 such examples to the prompt (see Appendix~\ref{app:icl-table}
for the full template) and rerun evaluation on \textbf{absence detection} across all fragmentation
rates.\footnote{Due to computational constraints, ICL evaluation is conducted
on Gemini 3.0 Flash on Legal Documents. Broader evaluation is left to future work.}

\begin{figure}[t]
  \centering
  \includegraphics[width=0.33\linewidth]{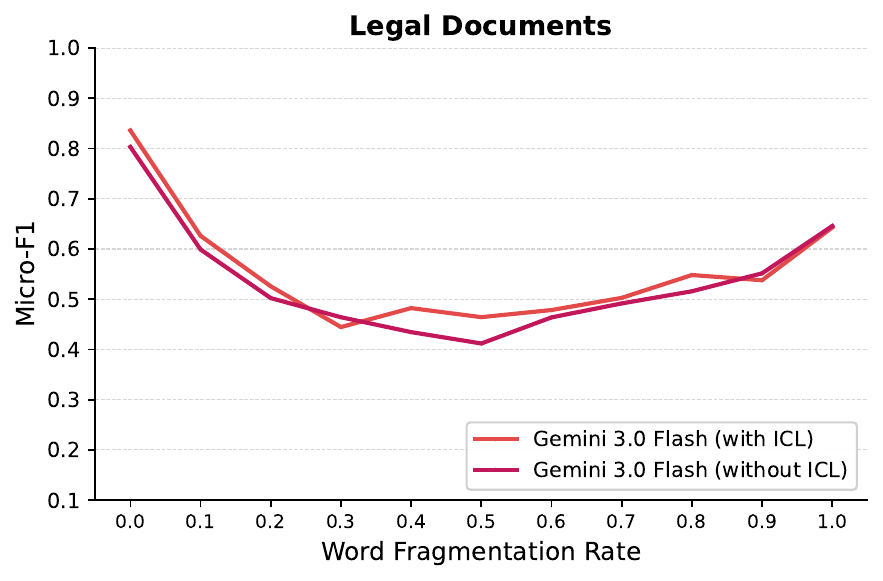}
  \caption{
    Effect of ICL on Micro-F1 across fragmentation rates for Gemini 3.0 Flash
    on legal documents. The U-shaped curve persists with and without ICL.
  }
  \label{fig:icl}
\end{figure}

\textbf{Results.} As shown in Figure~\ref{fig:icl}, the U-shaped curve persists after ICL with
shape and valley position largely unchanged.
ICL yields no significant performance improvement in most conditions and in
several conditions leads to a measurable performance decrease, consistent with
findings that corrupted demonstrations can mislead rather than guide models
\citep{gao2024iclnoise, jiao2026icldemo}.
The bottleneck is therefore not accessible to prompt level intervention,
suggesting that the valley arises from tokenization level processing rather
than from a lack of appropriate task strategy.

\begin{confirmed}[Finding 1: the bottleneck is below the prompting level]
ICL does not rescue valley-bottom performance. The valley's position and
shape are unchanged, ruling out task comprehension as the limiting factor.
\end{confirmed}

\subsection{Fixed-slot Fragmentation Substantially Reduces the U-shape}
\label{sec:exp-fixed}

\textbf{Setup.} We compare random-slot fragmentation against a \textbf{fixed-slot} variant that
inserts a space at the \emph{first slot only} of each contiguous non-space
character sequence with probability $p$; no spaces are inserted at any other
slot within the same sequence.\footnote{$p$ and \wfr{} are distinct: $p$ is the probability of inserting
a space after the first intra-word slot of each contiguous non-space character
sequence, while \wfr{} measures the fraction of \emph{all} intra-word slots
that received a space across the entire document; the same $p$ yields
different \wfr{} values depending on token-length distribution.}
This imposes a more regular, predictable insertion pattern.\footnote{Our codebase also supports inserting at the $n$-th slot or $m$-th
percentile slot of each sequence; we leave systematic evaluation of these variants to
future work.}

\begin{figure}[t]
  \centering
  \includegraphics[width=0.33\linewidth]{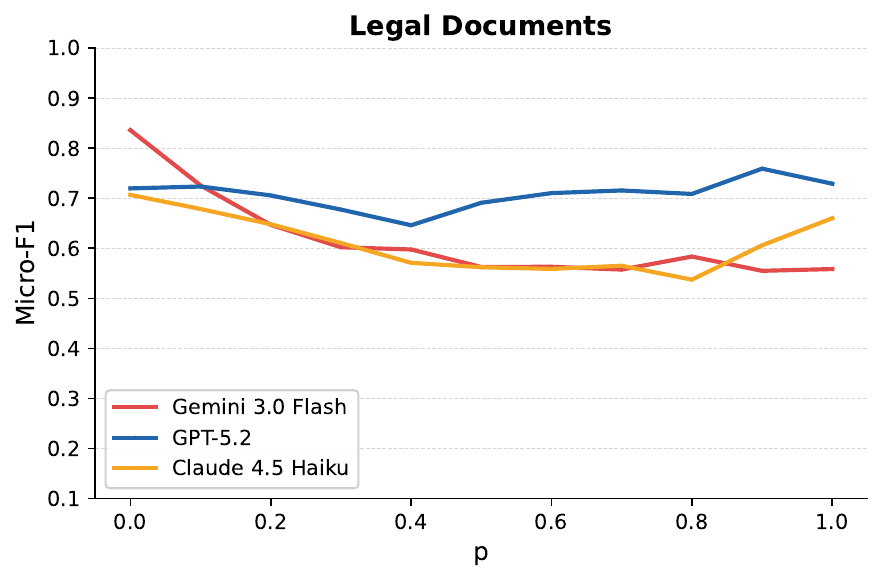}%
  \includegraphics[width=0.33\linewidth]{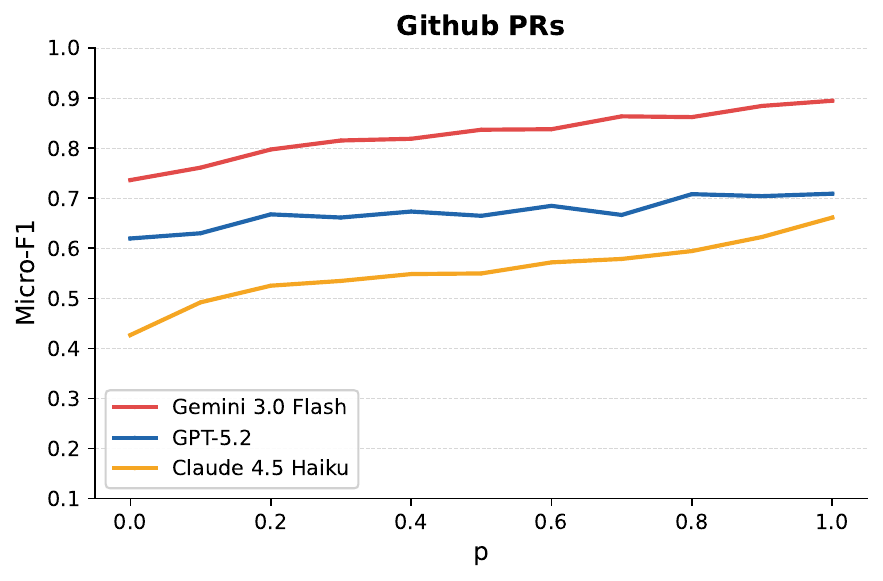}%
  \includegraphics[width=0.33\linewidth]{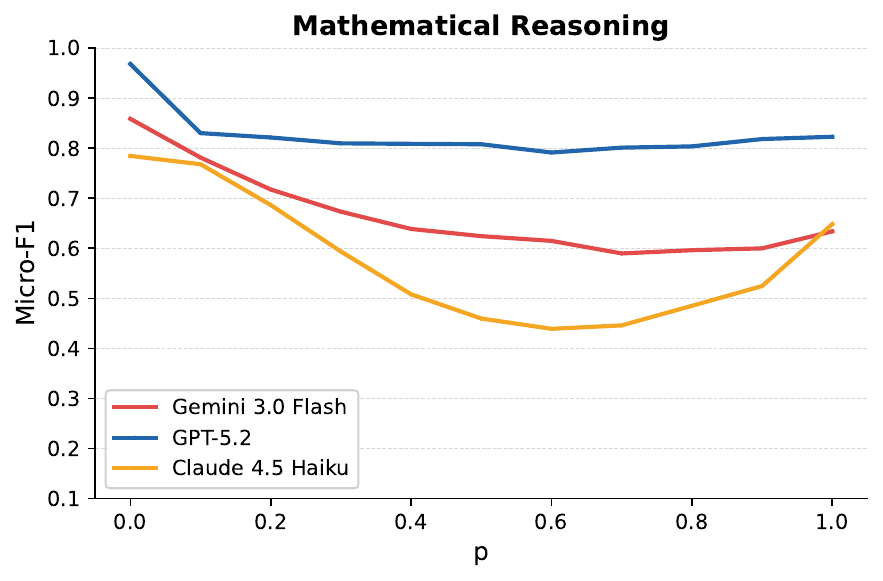}
  \caption{
    Fixed-slot fragmentation results across three domains.
    In the fixed-slot condition, a space is inserted at the first slot of each
    contiguous non-space character sequence with probability $p$.
    Each panel shows Micro-F1 as a function of $p$ for Gemini 3.0 Flash,
    GPT-5.2, and Claude 4.5 Haiku.
  }
  \label{fig:fixed}
\end{figure}

\textbf{Results on Legal Documents and Mathematical Reasoning.}
Fixed-slot fragmentation substantially reduces the U-shape on both
natural-language domains (Figure~\ref{fig:fixed}).
Gemini 3.0 Flash plateaus without a rebound; GPT-5.2 shows a similar pattern
on Mathematical Reasoning but no clear trend on Legal Documents.
Claude 4.5 Haiku still shows a partial uptick at high $p$, though less
pronounced than under random-slot.

\textbf{Results on GitHub PRs.}
Surprisingly, all three models show monotonically \emph{increasing} Micro-F1:
regular space insertion breaks dense identifier runs into shorter units,
consistently helping the model.

\textbf{Interpretation.}
These results support the two-modes hypothesis via two mechanisms.
First, regular fragmentation cannot fully activate the character-level regime at $p = 1.0$, so no rebound occurs. Second, it avoids the chaotic mixed tokenization state: the model adapts to a predictable structure rather than oscillating between regimes. The GitHub PRs inversion confirms that the valley is driven not by the amount of fragmentation, but by the disorder it introduces; regularizing this density eliminates the valley in natural language and actively assists the model on dense code diffs.
\begin{confirmed}[Finding 2: irregularity, not magnitude, drives the valley]
Regular fragmentation eliminates the rebound on natural-language domains and
\emph{monotonically improves} performance on code diffs. The valley arises
from \emph{how} text is fragmented, rather than \emph{how much}.
\end{confirmed}

\subsection{Inter-word Whitespace Manipulation Does Not Induce the Valley}
\label{sec:exp-control}

We conduct a control experiment where inter word spaces are randomly replaced with either two spaces or a space and a tab character.
Under this setup, performance remains largely unchanged and no U shaped curve emerges; this confirms that the valley arises from disrupting contiguous non space sequences, rather than from the mere act of inserting whitespace.

\begin{confirmed}[Finding 3: intra-word disruption is the cause]
Extra whitespace \emph{between} words does not produce the valley. Only
breaking apart intra-word character sequences, which disrupts subword
tokenization, triggers the effect.
\end{confirmed}

\subsection{Math Reasoning under Fragmentation}
\label{sec:exp-math}

To test whether the Text Uncanny Valley generalizes beyond absence detection and insert detection,
we apply the same fragmentation protocol to a direct math reasoning task
\citep{cobbe2021gsm8k}, giving each model individual problems at eleven
\wfr{} levels and evaluating accuracy
(see Appendix~\ref{app:math-reasoning} for full details).
Gemini 3.0 Flash retains a pronounced U-shaped curve, while GPT-5.2 and
Claude 4.5 Haiku remain near ceiling throughout.

\begin{figure}[t]
  \centering
  \includegraphics[width=0.33\linewidth]{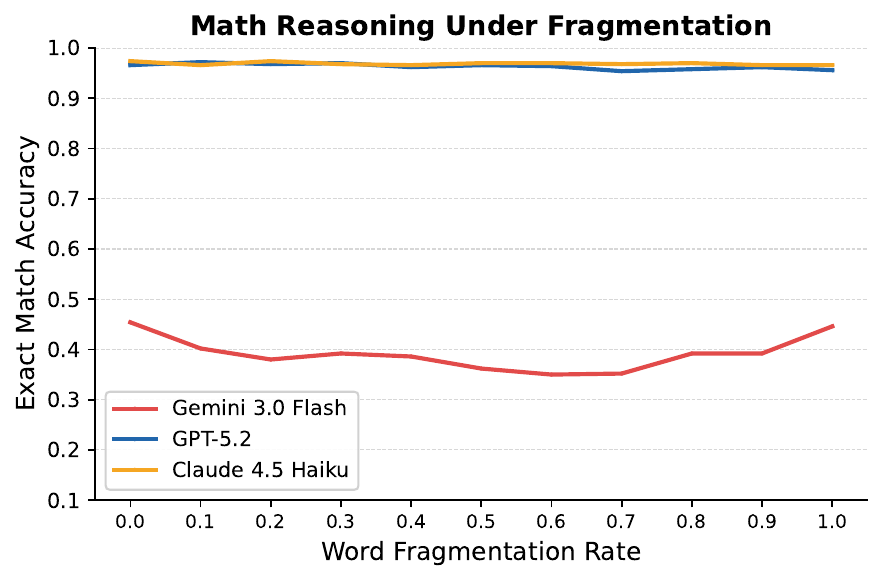}
  \caption{Math reasoning accuracy on individual GSM8K problems vs.\ \wfr{}.
  Thinking mode is disabled. Gemini 3.0 Flash shows a clear U-shape; GPT-5.2
  and Claude 4.5 Haiku remain near ceiling.}
  \label{fig:math-reasoning}
\end{figure}
Two factors likely explain the contrast: math reasoning does not require
exact cross-document lexical alignment, and both inputs and outputs are
substantially shorter than in the experiments in Section~\ref{sec:valley}, reducing the surface
area for tokenization
disruption to accumulate.
Both are consistent with the two-modes hypothesis: the U-shape is more
pronounced when a task requires the model to operate precisely in the
transition zone between word-level and character-level processing.

\begin{partialresult}[Finding 4: the U-shape is task and model dependent]
The valley persists for Gemini 3.0 Flash but disappears for GPT-5.2 and
Claude 4.5 Haiku on math reasoning.
The effect is strongest when exact cross-document alignment is required
over long inputs.
\end{partialresult}

\subsection{Analysis: Tokenization Entropy}
\label{sec:exp-entropy}

\textbf{Setup.}
We compute the Shannon entropy of the token-ID distribution for each document
at each \wfr{} level, using the \texttt{cl100k\_base} encoding accessed via \texttt{tiktoken}.\footnote{We verified that results
are nearly identical whether accessed as \texttt{encoding\_for\_model("gpt-4")} or
\texttt{get\_encoding("cl100k\_base")}.}
\begin{equation}
  H = -\sum_{t} p(t) \log_2 p(t)
  \label{eq:entropy}
\end{equation}
where $p(t)$ is the empirical frequency of token $t$ in the document.
High entropy indicates a diverse, heterogeneous token distribution; low entropy
indicates dominance by a small vocabulary (e.g., individual letters at high
fragmentation).
Shannon entropy is a standard measure of distributional diversity
\citep{shannon1948mathematical}.

\begin{figure}[t]
  \centering
  \includegraphics[width=0.33\linewidth]{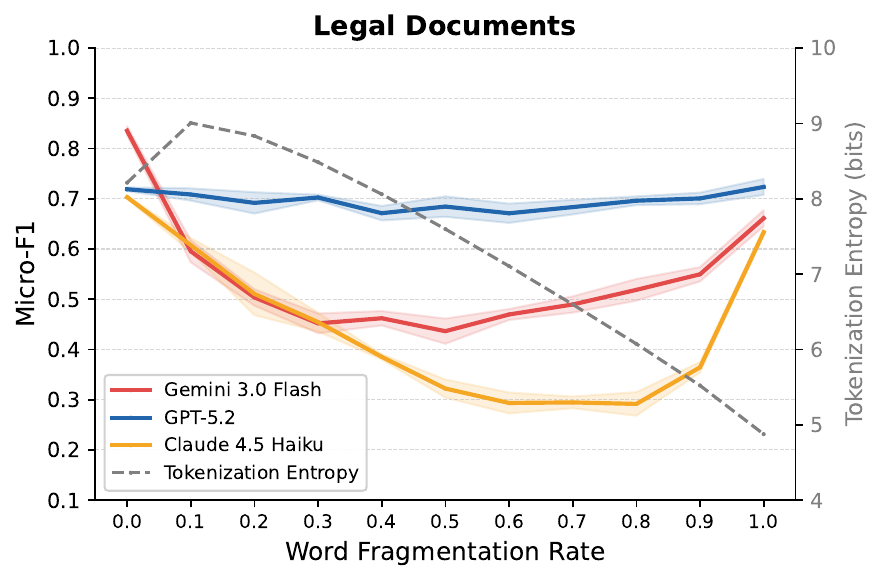}%
  \includegraphics[width=0.33\linewidth]{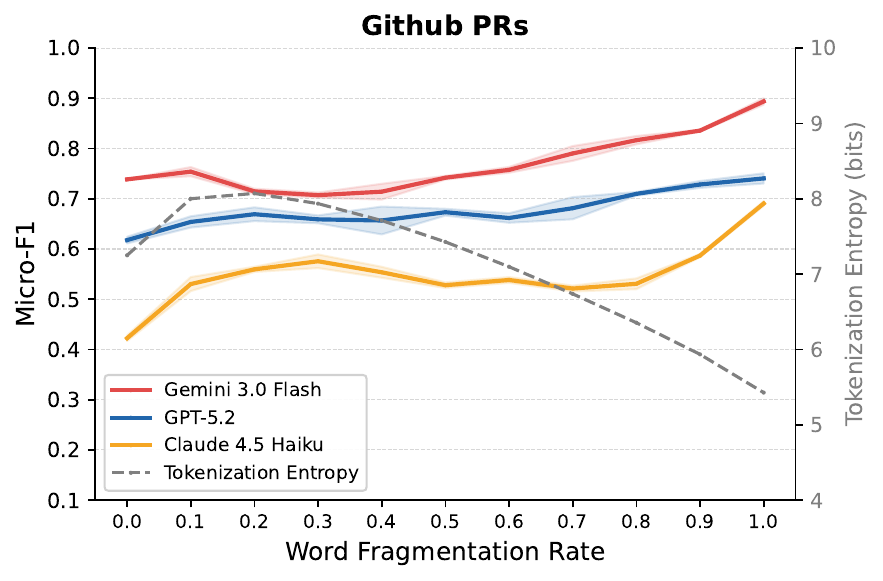}%
  \includegraphics[width=0.33\linewidth]{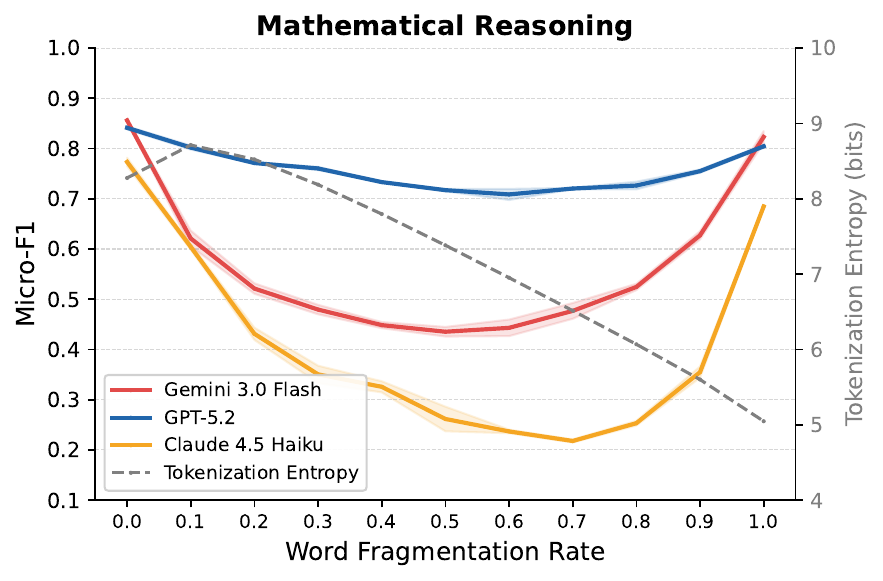}
  \caption{
    Tokenization entropy (dashed) overlaid on Micro-F1 (solid) from the absence
detection experiment, as a function
    of \wfr{} across three domains.
    Entropy first rises then falls as \wfr{} increases; the F1 curve follows
    the inverse shape.
  }
  \label{fig:entropy}
\end{figure}

\textbf{Results.}
As shown in Figure~\ref{fig:entropy}, tokenization entropy rises then falls as
\wfr{} increases across all three domains, forming an inverted U-shape that
mirrors the U-shaped F1 curve.
Notably, the entropy peak occurs at a lower \wfr{} than the F1 valley, i.e.,
entropy is already declining by the time performance reaches its minimum.

\textbf{Interpretation.}
The entropy peak occurs before the F1 minimum, indicating that maximum
tokenization heterogeneity alone does not account for the valley.
This offset is consistent with the two-modes hypothesis: the F1 minimum may correspond to a transition region where word-level processing has degraded while character-level processing has not yet stabilized, rather than to the point of maximum tokenization heterogeneity itself.
The entropy trajectory thus provides indirect support by showing that
the tokenization regime shifts non-monotonically across the fragmentation range.

\begin{confirmed}[Finding 5: the valley marks a regime transition]
Tokenization entropy peaks \emph{before} the F1 minimum across all three
domains. The valley is not the point of maximum token confusion but the
no-man's-land between two processing regimes.
\end{confirmed}
\section{Conclusion}
\label{sec:conclusion}

We document the \textbf{Text Uncanny Valley}: under word-boundary
fragmentation, LLM performance follows a U-shaped curve on absence detection
and NIAH-style insertion detection, reaching its minimum at moderate
fragmentation before partially recovering at extremes. This reveals a
clean-text-invisible failure mode, especially for tasks requiring exact lexical
alignment.

Our analyses support a two-regime account: ICL does not rescue the valley;
regularized fragmentation weakens it; inter-word whitespace does not induce it;
math reasoning shows the effect is model and task dependent; and tokenization
entropy peaks before the $F_1$ minimum. Together, these results suggest that
the valley marks an unstable transition between word-level and character-level
processing.

Moderate word-boundary corruption may therefore be more dangerous than extreme
fragmentation for OCR pipelines, document processing systems, and LLM-as-judge
evaluators. Future work should study how to
mitigate this transition-zone failure mode, and ensure robustness benchmarks
cover the ambiguous middle regime between clean and uniformly corrupted text.

\section*{Acknowledgements}

This work was supported by New Forms of Socio-Cognitive AI Initiative at the University of Chicago.

\bibliographystyle{abbrvnat}
\bibliography{references}

@inproceedings{fu2025absencebench,
  title     = {{Absence Bench}: Language Models Can't See What's Missing},
  author    = {Fu, Harvey Yiyun and Shrivastava, Aryan and Moore, Jared and
               West, Peter and Tan, Chenhao and Holtzman, Ari},
  booktitle = {The Thirty-ninth Annual Conference on Neural Information
               Processing Systems Datasets and Benchmarks Track},
  year      = {2025},
  url       = {https://openreview.net/forum?id=pmLMrqhIpb}
}

@inproceedings{jain2026sufficiency,
  title     = {Knowing What's Missing: Assessing Information Sufficiency
               in Question Answering},
  author    = {Jain, Akriti and Garimella, Aparna},
  booktitle = {Findings of the Association for Computational Linguistics: EACL 2026},
  pages     = {4163--4174},
  year      = {2026}
}

@inproceedings{ming2025faitheval,
  title     = {{FaithEval}: Can Your Language Model Stay Faithful to Context,
               Even If ``{The Moon is Made of Marshmallows}''?},
  author    = {Ming, Yifei and Purushwalkam, Senthil and Pandit, Shrey and
               Ke, Zixuan and Nguyen, Xuan-Phi and Xiong, Caiming and Joty, Shafiq},
  booktitle = {International Conference on Learning Representations},
  year      = {2025},
  url       = {https://openreview.net/forum?id=UeVx6L59fg}
}

@article{agrawal2025llmrobustness,
  title   = {Enhancing {LLM} Robustness to Perturbed Instructions: An Empirical Study},
  author  = {Agrawal, Aryan and Alazraki, Lisa and Honarvar, Shahin and Rei, Marek},
  journal = {arXiv preprint arXiv:2504.02733},
  year    = {2025},
  url     = {https://arxiv.org/abs/2504.02733}
}

@article{singh2024robustness,
  title   = {Robustness of Large Language Models to Perturbations in Text},
  author  = {Singh, Ayush and Singh, Navpreet and Vatsal, Shubham},
  journal = {arXiv preprint arXiv:2407.08989},
  year    = {2024},
  url     = {https://arxiv.org/abs/2407.08989}
}

@article{zhuo2025charperturb,
  title   = {On the Ability of {LLM}s to Handle Character-Level Perturbations:
             How Well and How?},
  author  = {Zhuo, Anyuan and Ning, Xuefei and Li, Ningyuan and Wang, Yu and Lu, Pinyan},
  journal = {arXiv preprint arXiv:2510.14365},
  year    = {2025},
  url     = {https://arxiv.org/abs/2510.14365}
}

@inproceedings{sclar2024prompt,
  title     = {Quantifying Language Models' Sensitivity to Spurious Features
               in Prompt Design or: How {I} Learned to Start Worrying about
               Prompt Formatting},
  author    = {Sclar, Melanie and Choi, Yejin and Tsvetkov, Yulia and Suhr, Alane},
  booktitle = {Proceedings of the International Conference on Learning
               Representations (ICLR)},
  year      = {2024},
  url       = {https://arxiv.org/abs/2310.11324}
}

@inproceedings{pawar2025brokenwords,
  title     = {Broken Words, Broken Performance: Effect of Tokenization on
               Performance of {LLM}s},
  author    = {Pawar, Sachin and Apte, Manoj and Jadhav, Kshitij and
               Palshikar, Girish Keshav and Ramrakhiyani, Nitin},
  booktitle = {Proceedings of the 14th International Joint Conference on Natural
               Language Processing and the 4th Conference of the Asia-Pacific
               Chapter of the Association for Computational Linguistics},
  pages     = {372--385},
  year      = {2025},
  url       = {https://aclanthology.org/2025.ijcnlp-short.31/}
}

@misc{kamradt2023niah,
  author       = {Kamradt, Greg},
  title        = {{LLMTest\_NeedleInAHaystack}: Pressure Testing {LLMs}},
  year         = {2023},
  howpublished = {\url{https://github.com/gkamradt/LLMTest_NeedleInAHaystack}}
}

@article{liu2024lostinmiddle,
  title   = {Lost in the Middle: How Language Models Use Long Contexts},
  author  = {Liu, Nelson F. and Lin, Kevin and Hewitt, John and Paranjape, Ashwin
             and Bevilacqua, Michele and Petroni, Fabio and Liang, Percy},
  journal = {Transactions of the Association for Computational Linguistics},
  volume  = {12},
  pages   = {157--173},
  year    = {2024}
}

@inproceedings{stop2026tokenizers,
  title     = {Stop Taking Tokenizers for Granted: They Are Core Design Decisions
               in Large Language Models},
  author    = {Alqahtani, Sawsan and Nayeem, Mir Tafseer and
               Laskar, Md Tahmid Rahman and Mohiuddin, Tasnim and Bari, M Saiful},
  booktitle = {Proceedings of the 19th Conference of the European Chapter of the
               Association for Computational Linguistics (Volume 1: Long Papers)},
  pages     = {8410--8432},
  year      = {2026},
  url       = {https://aclanthology.org/2026.eacl-long.394/}
}

@inproceedings{gao2024iclnoise,
  title     = {On the Noise Robustness of In-Context Learning for Text Generation},
  author    = {Gao, Hongfu and Zhang, Feipeng and Jiang, Wenyu and Shu, Jun
               and Zheng, Feng and Wei, Hongxin},
  booktitle = {Advances in Neural Information Processing Systems (NeurIPS)},
  volume    = {37},
  year      = {2024},
  url       = {https://arxiv.org/abs/2405.17264}
}

@article{jiao2026icldemo,
  title   = {Understanding the Dynamics of Demonstration Conflict in
             In-Context Learning},
  author  = {Jiao, Difan and Wang, Di and Hu, Lijie},
  journal = {arXiv preprint arXiv:2603.04464},
  year    = {2026}
}

@article{mori1970uncanny,
  title   = {Bukimi no Tani [{The Uncanny Valley}]},
  author  = {Mori, Masahiro},
  journal = {Energy},
  volume  = {7},
  number  = {4},
  pages   = {33--35},
  year    = {1970},
  note    = {In Japanese. English translation by Karl F. MacDorman and Norri Kageki,
             IEEE Robotics \& Automation Magazine, 19(2):98--100, 2012}
}

@article{seyama2007uncanny,
  title   = {The Uncanny Valley: Effect of Realism on the Impression of
             Artificial Human Faces},
  author  = {Seyama, Jun'ichiro and Nagayama, Ruth S.},
  journal = {Presence: Teleoperators and Virtual Environments},
  volume  = {16},
  number  = {4},
  pages   = {337--351},
  year    = {2007}
}

@article{cobbe2021gsm8k,
  title   = {Training Verifiers to Solve Math Word Problems},
  author  = {Cobbe, Karl and Kosaraju, Vineet and Bavarian, Mohammad and
             Chen, Mark and Jun, Heewoo and Kaiser, Lukasz and Plappert, Matthias
             and Tworek, Jerry and Hilton, Jacob and Nakano, Reiichiro and
             Hesse, Christopher and Schulman, John},
  journal = {arXiv preprint arXiv:2110.14168},
  year    = {2021},
  url     = {https://arxiv.org/abs/2110.14168}
}

@inproceedings{hendrycks2021cuad,
  title     = {{CUAD}: An Expert-Annotated {NLP} Dataset for Legal Contract Review},
  author    = {Hendrycks, Dan and Burns, Collin and Chen, Anya and Ball, Spencer},
  booktitle = {Proceedings of the Neural Information Processing Systems
               Track on Datasets and Benchmarks},
  year      = {2021},
  url       = {https://openreview.net/forum?id=7l1Ygs3Bamw}
}

@article{deepseekv3,
  title   = {{DeepSeek-V3} Technical Report},
  author  = {{DeepSeek-AI}},
  journal = {arXiv preprint arXiv:2412.19437},
  year    = {2024},
  url     = {https://arxiv.org/abs/2412.19437}
}

@misc{grok41,
  author       = {{xAI}},
  title        = {{Grok 4.1 Fast and Agent Tools API}},
  year         = {2025},
  howpublished = {\url{https://x.ai/news/grok-4-1-fast}}
}

@article{qwen3,
  title   = {{Qwen3} Technical Report},
  author  = {Yang, An and Li, Anfeng and Yang, Baosong and others},
  journal = {arXiv preprint arXiv:2505.09388},
  year    = {2025},
  url     = {https://arxiv.org/abs/2505.09388}
}

@misc{gemini3flash,
  author       = {{Google}},
  title        = {Gemini 3 Flash},
  year         = {2025},
  howpublished = {\url{https://cloud.google.com/blog/products/ai-machine-learning/gemini-3-flash-for-enterprises}}
}

@misc{claudehaiku45,
  author       = {{Anthropic}},
  title        = {System Card: {Claude Haiku 4.5}},
  year         = {2025},
  howpublished = {\url{https://anthropic.com/claude-haiku-4-5-system-card}}
}

@article{shannon1948mathematical,
  title   = {A Mathematical Theory of Communication},
  author  = {Shannon, Claude E.},
  journal = {The Bell System Technical Journal},
  volume  = {27},
  number  = {3},
  pages   = {379--423},
  year    = {1948}
}

@article{yang2026wordrecovery,
  title   = {Word Recovery in Large Language Models Enables Character-Level
             Tokenization Robustness},
  author  = {Yang, Zhipeng and Yang, Shu and Hu, Lijie and Wang, Di},
  journal = {arXiv preprint arXiv:2603.10771},
  year    = {2026},
  url     = {https://arxiv.org/abs/2603.10771}
}

@inproceedings{chai2024tokenizationfalling,
  title     = {Tokenization Falling Short: On Subword Robustness in Large
               Language Models},
  author    = {Chai, Yekun and Fang, Yewei and Peng, Qiwei and Li, Xuhong},
  booktitle = {Findings of the Association for Computational Linguistics: EMNLP 2024},
  pages     = {1582--1599},
  year      = {2024},
  url       = {https://aclanthology.org/2024.findings-emnlp.86/}
}

@article{piryani2025multiocr,
  title   = {{MultiOCR-QA}: Dataset for Evaluating Robustness of {LLM}s in
             Question Answering on Multilingual {OCR} Texts},
  author  = {Piryani, Bhawna and Mozafari, Jamshid and Abdallah, Abdelrahman
             and Doucet, Antoine and Jatowt, Adam},
  journal = {arXiv preprint arXiv:2502.16781},
  year    = {2025},
  url     = {https://arxiv.org/abs/2502.16781}
}

@inproceedings{sennrich2016bpe,
  title     = {Neural Machine Translation of Rare Words with Subword Units},
  author    = {Sennrich, Rico and Haddow, Barry and Birch, Alexandra},
  booktitle = {Proceedings of the 54th Annual Meeting of the Association
               for Computational Linguistics (Volume 1: Long Papers)},
  pages     = {1715--1725},
  year      = {2016},
  url       = {https://aclanthology.org/P16-1162/}
}

@article{hsieh2024ruler,
  title   = {{RULER}: What's the Real Context Size of Your Long-Context
             Language Models?},
  author  = {Hsieh, Cheng-Ping and Sun, Simeng and Kriman, Samuel and
             Acharya, Shantanu and Rekesh, Dima and Jia, Fei and
             Zhang, Yang and Ginsburg, Boris},
  journal = {arXiv preprint arXiv:2404.06654},
  year    = {2024},
  url     = {https://arxiv.org/abs/2404.06654}
}

@misc{gpt52systemcard,
  author       = {{OpenAI}},
  title        = {Update to {GPT-5} System Card: {GPT-5.2}},
  year         = {2025},
  howpublished = {\url{https://openai.com/index/gpt-5-system-card-update-gpt-5-2/}}
}

\newpage
\appendix

\section{Limitations}
\label{sec:limitations}

\paragraph{Indirect mechanism evidence.}
Our two-modes hypothesis rests on behavioral observations and tokenization
statistics rather than internal model states.
A direct mechanistic account would require access to attention patterns or
layer-wise activations, none of which are available through black-box APIs.
The entropy analysis provides indirect support that is consistent with the
hypothesis but does not confirm it causally; future work with open-weight
models could probe internal representations directly.

\paragraph{Model coverage and proprietary models.}
Only three of eight evaluated model families exhibit a clear U-shape.
We cannot determine whether the others fail to do so due to training data
composition, architectural choices, tokenizer design, or safety-related
output filtering.
All evaluated models are proprietary, which limits reproducibility and
makes it impossible to control for training data.
Replication with open-weight models of documented training composition
(e.g., LLaMA, Mistral, or Qwen open variants) would allow more controlled
comparisons and is an important direction for future work.

\paragraph{Domain and task coverage.}
Although we evaluate word-boundary fragmentation across absence detection,
NIAH-style insertion detection, and a direct math reasoning task, our coverage
remains limited. The strongest evidence comes from long-context
document-comparison settings, especially absence detection, where exact lexical
alignment is central to the task. Our math reasoning experiment extends beyond
document comparison, but only to GSM8K-style problems and shows that the effect
can be model- and task-dependent. Whether the Text Uncanny Valley appears in
other task types, such as question answering, summarization, reading
comprehension, or code generation, remains unknown. These tasks impose
different forms of lexical alignment pressure and may exhibit different
sensitivity to word-boundary corruption. Similarly, we fix the omission and
insertion rates at $p = 0.1$ for the document-comparison tasks; although we
verified that varying this rate does not substantially change the U-shaped
curve, a systematic study across target rates is beyond the scope of this work.

\paragraph{Perturbation scope.}
We study only intra-word space insertion as the corruption mechanism.
Real-world OCR errors include character substitution, deletion, transposition,
and merging, which may interact differently with absence detection and may or
may not produce non-monotonic performance curves.
Our analysis is also conducted at the whitespace-delimited word level rather
than the subword token level; an intra-token analysis studying how fragmentation
interacts with tokenizer boundaries directly could yield finer-grained
mechanistic insight and is a promising direction for future work.
The thinking-mode evaluation is additionally limited to two model families on
one domain, so the observed asymmetry between GPT-5.2 and Gemini 3.0 Flash
should not be generalized without broader investigation.

\paragraph{Generalisability beyond absence detection.}
We evaluate word-boundary fragmentation beyond absence detection through a
NIAH-style insertion detection task (Appendix~\ref{app:niah}) and a direct
GSM8K math reasoning task (Appendix~\ref{app:math-reasoning}). The NIAH-style
task shows a weaker but still observable U-shaped effect, consistent with our
argument that absence detection is a particularly sensitive testbed because it
requires exact lexical alignment between two document versions. The math
reasoning task further suggests that the effect is model- and task-dependent:
Gemini 3.0 Flash exhibits a clear U-shape, while GPT-5.2 and Claude 4.5 Haiku
remain near ceiling. These results extend the phenomenon beyond absence
detection, but they do not establish full generality. Whether the Text Uncanny
Valley appears in other task types, such as question answering, summarization,
reading comprehension, or code generation, remains unknown. Future work should
test whether the U-shaped curve generalizes to broader long-context alignment
and text-processing tasks.

\paragraph{Statistical power and computational cost.}
We evaluate 100 documents per domain with three fragmentation seeds.
While sufficient to reveal consistent U-shaped trends, this limits
the statistical power of fine-grained analyzes such as precisely locating
the valley bottom or quantifying its depth across document subsets.

Scaling to larger samples was not feasible given the structure of our
evaluation protocol.
Each document requires 11 separate API calls, one per \wfr{} level, and
each prompt is approximately twice the length of the original document,
because the \ab{} task format requires both the original and modified
contexts in the same prompt.
Our documents average several thousand tokens in length; across 100 documents,
11 \wfr{} levels, 3 model families, 3 domains, and 3 fragmentation seeds,
the main experiments alone require approximately 29,700 API calls
($100 \times 11 \times 3 \times 3 \times 3$), not counting NIAH, ICL,
fixed-slot, and thinking-mode variants, which add further thousands.
Many documents, especially in the legal domain, are long enough to require
model context windows of 32K tokens or more.
The combined effect of long prompts, many API calls per document, and high
per-call latency made the full evaluation both time-consuming and costly,
and was conducted at the limit of our available computational resources.

\section{Math Reasoning Under Fragmentation}
\label{app:math-reasoning}

We apply the same random-slot fragmentation protocol
(Algorithm~\ref{alg:random-slot}) to individual GSM8K problems to test whether
the Text Uncanny Valley generalizes beyond absence detection.

\paragraph{Setup.}
We randomly sample 500 problems from the GSM8K test split (seed 42) and
evaluate each model on each problem independently at eleven \wfr{} levels
($0.0, 0.1, \ldots, 1.0$).
Each API call consists of a single fragmented problem; thinking mode is
disabled for all models to isolate tokenization level effects.
Accuracy is measured by exact numerical match, with tolerance for minor
formatting differences (e.g., $13.00 = 13$), using the pattern
\texttt{The answer is \#\#\#\# [result]} as the primary extraction target,
with fallback to the last number in the response.
Both the input context (a single problem) and the expected output (a single
number) are substantially shorter than in the absence detection experiments,
which use 80-problem concatenated documents.

\paragraph{Prompt.}
Each model receives the following prompt, where \texttt{\{question\}} is replaced
with the fragmented problem text:

\begin{table}[H]
\centering
\small
\newcommand{\phmath}[1]{\textcolor{blue}{\texttt{\{#1\}}}}
\caption{Prompt template used in the math reasoning experiment.
The placeholder \textcolor{blue}{\texttt{\{question\}}} is replaced with the
fragmented problem text at each \wfr{} level.}
\label{tab:math-prompt}
\begin{tabular}{@{}p{13cm}@{}}
\toprule
\textbf{User Message} \\
\midrule
\texttt{Question: \phmath{question}}\\[4pt]
\texttt{Solve the problem step by step and end with}\\
\texttt{'The answer is \#\#\#\# [result]'.}\\
\bottomrule
\end{tabular}

\end{table}

\section{NIAH Results Under Fragmentation}
\label{app:niah}

We evaluate word-boundary fragmentation on a NIAH-style task using the same
\ab{} framework \citep{fu2025absencebench}, adapted to detect \emph{added}
rather than missing lines.
Specifically, each instance presents the model with both the original document
and a modified version to which extra lines have been inserted; the model must
identify exactly which lines were added, scored by Micro-F1.
The prompt template is provided below (Table~\ref{tab:niah-prompt}).

\begin{table}[H]
\centering
\small
\newcommand{\phb}[1]{\textcolor{blue}{\texttt{\{#1\}}}}
\caption{Prompt template used in the NIAH control experiment
(Appendix~\ref{app:niah}). Unlike the main prompt (Table~\ref{tab:prompt}),
this variant asks the model to identify \emph{added} lines rather than
\emph{missing} ones. No ICL prefix is used.}
\label{tab:niah-prompt}
\begin{tabular}{@{}p{13cm}@{}}
\toprule
\textbf{System Prompt} \\
\midrule
\texttt{You are helping someone identify components that were added to a
\phb{document\_type}.
The person will provide their version of the \phb{document\_type}, but they
may have added some random lines that were not in the original text.
Your task is to identify exactly which lines are not in the original
\phb{document\_type}. List only the extra lines, nothing else.} \\
\midrule
\textbf{User Message} \\
\midrule
\texttt{Here is the complete original \phb{document\_type}:}\\
\texttt{\phb{original\_context}}\\[2pt]
\texttt{Now, here is my recitation with some extra lines that were not in
the original \phb{document\_type}:}\\
\texttt{\phb{modified\_context}}\\[2pt]
\texttt{What lines did I add to the \phb{document\_type}? Please list only
the extra lines, nothing else.}\\
\bottomrule
\end{tabular}

\end{table}

We apply the same eleven \wfr{} levels across all three model families and
three domains.
The U-shaped effect is substantially weaker than in absence detection.
On GitHub PRs, all three models remain near-ceiling with no discernible U-shape.
On legal documents, mild non-monotonic variation is visible but not pronounced.
On mathematical reasoning, Claude 4.5 Haiku shows a more visible U-shape (from
approximately 0.95 to 0.75 and back), while GPT-5.2 and Gemini 3.0 Flash remain
largely stable.
Overall, the effect is attenuated compared to absence detection, consistent with
our argument that absence detection is a more sensitive testbed: it operates far
from ceiling and requires exact lexical alignment across two document versions,
amplifying the impact of tokenization disruption.

\paragraph{Needle design.}
For each domain, we construct 100 needle sentences whose length distribution
matches the per-line length distribution of the corresponding dataset.
The needles are domain-appropriate text that does not appear in the original
datasets: synthetic legal clauses for legal documents, synthetic code diff
hunks for GitHub PRs, and synthetic math word problems for GSM8K.
For each document, we sample a set of synthetic needle lines so that the number of inserted lines matches the insertion rate used in the main experiments. Needles are sampled from a domain-specific pool and inserted at random positions.
Crucially, the inserted needle is subjected to the same \wfr{} fragmentation
as the surrounding document, ensuring that the model must detect an added
fragment-level span rather than a clean insertion.

Table~\ref{tab:niah-examples} shows representative needle examples at their
respective fragmentation levels.

\begin{table}[H]
\centering
\small
\caption{Representative needle examples at domain-specific \wfr{} levels.
Needles are domain-appropriate synthetic text, subjected to the same
fragmentation as the surrounding document.}
\label{tab:niah-examples}
\begin{tabular}{@{}p{2.2cm}p{10.5cm}@{}}
\toprule
\textbf{Domain (\wfr{})} & \textbf{Needle example} \\
\midrule
GitHub PRs ($0.3$) &
\texttt{dif f - -git a/src / Illum in ate/A uth /Acce s s / G at e .p hp b/src/...
@@ -1 42,6 +154,12 @@ p rot ect ed fu n ction r esolveAuthC allba ck...
-        r e turn \$ thi s->co ntaine r->make(\$c a l lba ck)- > h a n d l e(\$user , ...\$a rgu men ts);} \\[4pt]
Legal Docs ($0.5$) &
\texttt{Not w ith s ta n d i ng t h e p ro visi o ns of S ect i o n 4 .2 ,
Va n te d g e C a pi t a l sh a l l r em i t a q ua r terl y i n frastru ct u r e
o pti m izat i on pa y m e nt o f \$ 1 8, 4 5 0 t o th e p r ovid er [...]
T h is M a s te r A g ree m e n t s h al l b e a ut om a t i call y ex tend e d
f o r a n ad di ti o n al ni n e -m on t h t erm u nles s [...]
} \\[4pt]
Math ($0.7$) &
\texttt{S a r a h h as a c o l le c ti on o f 5 0 v in t age p os t a g e s t a m p s .
S h e d eci de s t o s e l l 12 of t he m at a n a u c t i o n to buy 8 r a re on e s .
H er gra n df a the r t h e n g ift s h e r h is ow n co l l e c ti on w h i c h is thre e
tim e s t h e si ze o f Sa rah ' s c u r r e n t set .
H ow m any s tam ps d o e s s he h a v e i n t o t al n o w ?} \\
\bottomrule
\end{tabular}

\end{table}

\section{Results for All Evaluated Models}
\label{app:all-models}

We evaluated eight model variants in total.
As noted in Section~\ref{sec:valley-main}, only Gemini 3.0 Flash, GPT-5.2,
and Claude 4.5 Haiku consistently exhibit the Text Uncanny Valley.
This appendix provides the full Micro-F1 curves for the remaining variants.

\begin{figure}[H]
  \centering
  \includegraphics[width=0.33\linewidth]{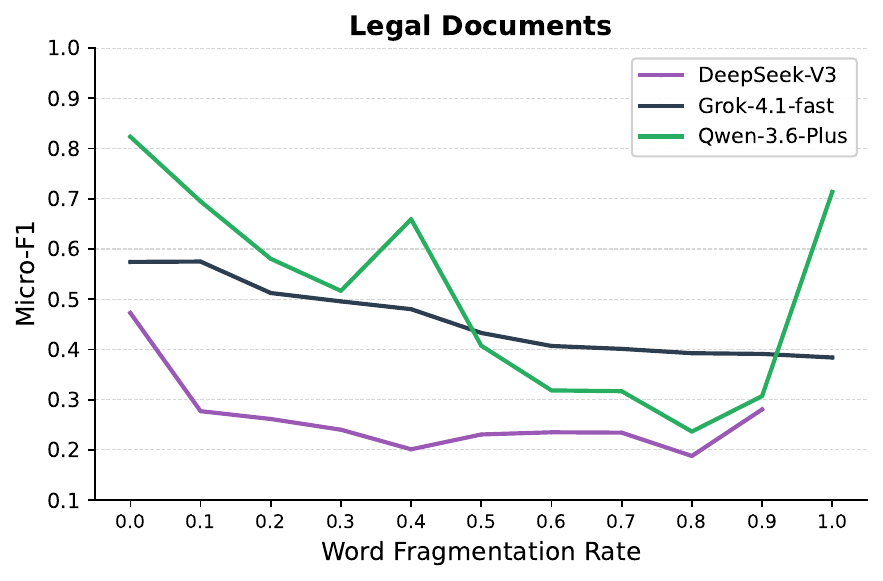}
  \caption{Micro-F1 as a function of \wfr{} for DeepSeek-V3, Grok-4.1-Fast,
  and Qwen3-Plus on legal documents, with all three models overlaid.
  DeepSeek-V3 results at \wfr{}~$= 1.0$ are omitted due to consistent API
  errors at that fragmentation level.}
  \label{fig:all-models-ld}
\end{figure}

\paragraph{Non-U-shape models (legal documents).}
Figure~\ref{fig:all-models-ld} shows results for DeepSeek-V3, Grok-4.1-Fast,
and Qwen3-Plus on legal documents.
Grok-4.1-Fast shows largely monotonic decline without a clear rebound.
Qwen3-Plus exhibits irregular curve shapes with no consistent trend across
fragmentation rates; its overall performance is poor, and API calls were
substantially slower and more costly than other models, which precluded its
inclusion in further experiments.
DeepSeek-V3 exhibits an important anomaly: as \wfr{} increases, certain
documents consistently trigger internal errors or refusals, and at \wfr{}~$=1.0$
the model is almost entirely non-functional on this domain.
We conjecture this reflects an internal content-filtering mechanism that is
triggered by the unusual character distribution of fully fragmented text.
Due to these errors, we omit the \wfr{}~$= 1.0$ data point for DeepSeek-V3
on legal documents from all figures and analyzes.

\section{Experimental Details}
\label{app:details}

\subsection{GSM8K document structure}
\label{app:gsm8k-table}

\begin{table}[h]
\centering
\small
\caption{Structure of a GSM8K document. Each document concatenates 80
problems; questions and answers are each split into individual lines.}
\label{tab:gsm8k-structure}
\begin{tabular}{ll}
\toprule
\textbf{Line index} & \textbf{Content} \\
\midrule
1       & Problem 1, question line 1 \\
\ldots  & \ldots \\
$k$     & Problem 1, question line $k$ \\
$k{+}1$ & Problem 1, answer line 1 \\
\ldots  & \ldots \\
$N$     & Problem 80, answer line (last) \\
\bottomrule
\end{tabular}

\end{table}

\subsection{Prompt template}
\label{app:prompt-table}

\begin{table}[H]
\centering
\small
\newcommand{\ph}[1]{\textcolor{blue}{\texttt{\{#1\}}}}
\caption{Prompt template used in all experiments. Placeholders in
\textcolor{blue}{blue} are filled at runtime.
In our experiments, \ph{document\_type} is replaced with
\texttt{Legal Documents}, \texttt{Github PRs}, or \texttt{Mathematical Reasoning}
depending on the domain.
The ICL prefix is prepended only when ICL is enabled
(Section~\ref{sec:exp-icl}).}
\label{tab:prompt}
\begin{tabular}{@{}p{13cm}@{}}
\toprule
\textbf{System Prompt} \\
\midrule
\texttt{You are helping someone identify missing components from a
\ph{document\_type}.
The person will provide their version of the \ph{document\_type}, but they
may have missed some lines. Your task is to identify exactly which lines are
missing from their version. List only the missing lines, nothing else.} \\
\midrule
\textbf{User Message} \\
\midrule
\textit{(Optional ICL prefix:)} \texttt{Below are some examples of the task:
\ph{icl\_examples} ---} \\[4pt]
\texttt{Here is the complete original \ph{document\_type}: \ph{original\_context}}\\[2pt]
\texttt{Now, here is my recitation which may be missing some lines:
\ph{modified\_context}}\\[2pt]
\texttt{What lines did I miss? Please list only the missing lines, nothing else.}\\
\bottomrule
\end{tabular}

\end{table}

\subsection{Fragmentation examples}
\label{app:frag-table}

\begin{table}[H]
\centering
\small
\caption{Effect of word-boundary fragmentation at four \wfr{} levels on a
legal document excerpt. Only intra-word slots receive inserted spaces;
inter-word spaces and punctuation are unchanged.}
\label{tab:fragmentation-examples}
\begin{tabular}{@{}lp{9.5cm}@{}}
\toprule
\textbf{\wfr{}} & \textbf{Text excerpt} \\
\midrule
0.0 &
This Promotion Agreement (the ``Agreement'') is dated as of January 26, 1998
between CNET, Inc.\ (``CNET'') and Cyberian Outpost, Inc.\ (the ``Company'').\ [\ldots] \\[4pt]
0.1 &
This Promoti on A greemen t (the ``Agreement'') is dated as of January 26, 1998
between CNET, I nc.\ (``CN ET'') and Cyberian Ou tpost, Inc.\ (the ``Company'').\ [\ldots] \\[4pt]
0.5 &
T h is P r om oti o n A greem e n t (th e ``A g re e m e n t'') is dated a s of
J an ua r y 26, 1998 b e twe en CN ET, I nc.\ (``CN ET'') a n d Cy be r i an
O u t p o st, I n c.\ (th e ``Com p an y'').\ [\ldots] \\[4pt]
1.0 &
T h i s P r o m o t i o n A g r e e m e n t ( t h e ``A g r e e m e n t'')
i s d a t e d a s o f J a n u a r y 2 6 , 1 9 9 8 b e t w e e n C N E T ,
I n c .\ (``C N E T'') a n d C y b e r i a n O u t p o s t , I n c .
( t h e ``C o m p a n y'').\ [\ldots] \\
\bottomrule
\end{tabular}

\end{table}

\subsection{ICL example template}
\label{app:icl-table}

\begin{table}[H]
\centering
\small
\caption{ICL prefix prepended to the user message. Each example shows a
fragmented document, a modified version with lines removed, and the correct
answer. Only Example 1 is shown in full.}
\label{tab:icl-example}
\begin{tabular}{@{}p{13cm}@{}}
\toprule
\textbf{ICL prefix prepended to User Message} \\
\midrule
\texttt{Below are some examples of the task:} \\[4pt]
\texttt{Example 1:}\\
\texttt{Here is the complete original document:}\\
\texttt{Th e fund amen tal pr incipl es of qu ant um co mpu ting inv olve s uperpo sition .}\\
\texttt{Th ese p henomen a allow f or com putatio nal spe eds f ar excee ding cl assic al ar chitectu re .}\\
\texttt{Imple ment ing r obust er ror co rrecti on rem ains a s ignific ant h urdle f or res earchers .}\\
\texttt{Cr yogen ic en vironmen ts a re nece ssary t o ma intain t he stabili ty o f su percon ducting q ubits .}\\
\texttt{[\ldots]}\\[4pt]
\texttt{Now, here is my recitation which may be missing some lines:}\\
\texttt{Th e fund amen tal pr incipl es of qu ant um co mpu ting inv olve s uperpo sition .}\\
\texttt{Th ese p henomen a allow f or com putatio nal spe eds f ar excee ding cl assic al ar chite ctu re .}\\
\texttt{Cr yogen ic en vironmen ts a re nece ssary t o ma intain t he stabili ty o f su percon ducting q ubits .}\\
\texttt{[\ldots]}\\[4pt]
\texttt{What lines did I miss? Please list only the missing lines, nothing else.}\\
\texttt{Answer:}\\
\texttt{Imple ment ing r obust er ror co rrecti on rem ains a s ignific ant h urdle f or res earchers .}\\[4pt]
\texttt{---}\\
\texttt{Example 2: [\ldots]}\\
\texttt{---}\\
\texttt{Example 3: [\ldots]}\\
\midrule
\texttt{---}\\
\textit{(Standard user message follows here.)}\\
\bottomrule
\end{tabular}

\end{table}

\subsection{Fragmentation procedure}

Algorithm~\ref{alg:random-slot} describes the random-slot fragmentation used
in the main experiments.
Algorithm~\ref{alg:fixed-slot} describes the fixed-slot variant used in
Experiment~B (Section~\ref{sec:exp-fixed}).

\begin{algorithm}[H]
\caption{Random-slot fragmentation}
\label{alg:random-slot}
\begin{algorithmic}[1]
\Require Document lines $L$, insertion probability $p \in [0,1]$
\Ensure Fragmented lines $L'$, measured $\wfr \in [0,1]$
\State $\text{inserted} \gets 0$;\ $\text{eligible} \gets 0$
\For{each line $\ell \in L$}
  \State $\ell' \gets \texttt{""}$
  \For{$i = 0$ \textbf{to} $|\ell| - 1$}
    \State $\ell' \gets \ell' + \ell[i]$
    \If{$i+1 < |\ell|$ \textbf{and} $\ell[i] \notin \text{whitespace}$ \textbf{and} $\ell[i+1] \notin \text{whitespace}$}
      \State $\text{eligible} \gets \text{eligible} + 1$
      \If{$\text{Bernoulli}(p) = 1$}
        \State $\ell' \gets \ell' + \texttt{" "}$;\ $\text{inserted} \gets \text{inserted} + 1$
      \EndIf
    \EndIf
  \EndFor
  \State Append $\ell'$ to $L'$
\EndFor
\State \Return $L'$,\ $\wfr = \text{inserted} / \max(\text{eligible}, 1)$
\end{algorithmic}
\end{algorithm}

\begin{algorithm}[H]
\caption{Fixed-slot fragmentation (first-slot variant)}
\label{alg:fixed-slot}
\begin{algorithmic}[1]
\Require Document lines $L$, insertion probability $p \in [0,1]$
\Ensure Fragmented lines $L'$, measured rate
\State $\text{inserted} \gets 0$;\ $\text{eligible} \gets 0$
\For{each line $\ell \in L$}
  \State Split $\ell$ into non-space sequences $S = [s_1, s_2, \ldots]$ and whitespace tokens
  \For{each sequence $s \in S$}
    \If{$|s| \geq 2$}
      \State $\text{eligible} \gets \text{eligible} + (|s| - 1)$ \Comment{count all gaps in sequence}
      \If{$\text{Bernoulli}(p) = 1$}
        \State Insert \texttt{" "} after $s[0]$ \Comment{first slot only}
        \State $\text{inserted} \gets \text{inserted} + 1$
      \EndIf
    \EndIf
  \EndFor
  \State Reconstruct $\ell'$ from modified sequences and original whitespace
  \State Append $\ell'$ to $L'$
\EndFor
\State \Return $L'$,\ $\text{rate} = \text{inserted} / \max(\text{eligible}, 1)$
\end{algorithmic}
\end{algorithm}

\end{document}